\documentclass{llncs}

\usepackage{makeidx}  % allows for indexgeneration
\usepackage{etoolbox} % for "\patchcmd" macro
\makeatletter
\patchcmd{\ps@headings}{\rlap{\thepage}}{}{}{}
\patchcmd{\ps@headings}{\llap{\thepage}}{}{}{}
\makeatother
\usepackage{xcolor}
\usepackage{array}
\usepackage[caption=false,font=normalsize,labelfont=sf,textfont=sf]{subfig}
\usepackage{textcomp}
\usepackage{stfloats}
\usepackage{url}
\usepackage{verbatim}
\usepackage{graphicx}
\usepackage{cite}
\usepackage{algpseudocode,algorithm,algorithmicx}

\usepackage{amsmath,amssymb,amsfonts}
\usepackage{optidef}
\usepackage{textcomp}
\usepackage{multirow}
\usepackage{romannum}
\usepackage{diagbox}
\usepackage{enumitem}
\usepackage{bm}
\usepackage[hidelinks]{hyperref}

\begin{document}

\mainmatter 
%\title{Aggressive Trajectory Tracking of Quadcopter Using Embedded Model Predictive Control}
\title{{Aggressive Trajectory Tracking for Nano Quadrotors Using Embedded Nonlinear Model Predictive Control}}

%\author{Muhammad Kazim$^{1}$, Kwang-Ki K. Kim$^{2*}$, Hyunjae Sim$^{3}$, Gihun Shin$^{4}$
% \author{Muhammad Kazim, Hyunjae Sim, Gihun Shin and Kwang-Ki K. Kim

%
\author{Muhammad Kazim\inst{1} \and Hyunjae Sim\inst{1} Gihun Shin\inst{1} Hwancheol Hwang\inst{1} \and Kwang-Ki K. Kim\inst{1,*}}
\institute{
Department of Electrical and Computer Engineering \\ Inha University, Republic of Korea 
%\and
%Department of Electrical Engineering, Inha University, Republic of Korea 
\\[1mm]
$^{*}$Corresponding author: K.-K.~K. Kim ({\tt kwangki.kim@inha.ac.kr})}

\maketitle

\begin{abstract}

This paper presents an aggressive trajectory tracking method for a small lightweight nano-quadrotor using nonlinear model predictive control (NMPC) based on {\tt acados}. Controlling a nano quadrotor for accurate trajectory tracking at high speed in dynamic environments is challenging due to complex aerodynamic forces that introduce significant disturbances and large positional tracking errors. These aerodynamic effects are difficult to be identified and require feedback control that compensates for them in real time. NMPC allows the nano-quadrotor to control its motion in real time based on onboard sensor measurements, making it well-suited for tasks such as aggressive maneuvers and navigation in complex and dynamic environments. The software package {\tt acados} enables the implementation of the NMPC algorithm on embedded systems, which is particularly important for nano-quadrotor due to its limited computational resources. Our autonomous navigation system is developed based on an AI-deck that is a GAP8-based parallel ultra-low power computing platform with onboard sensors of a multi-ranger deck and a flow deck. The proposed method of NMPC-based trajectory tracking control is tested in simulation and the results demonstrate its effectiveness in trajectory tracking while considering the dynamic environments. It is also tested on a real nano quadrotor hardware, 27-g Crazyflie 2.1, with a customized MCU running embedded NMPC, in which accurate trajectory tracking results are achieved in dynamic real-world environments.

\keywords{Optimal tracking control, Nonlinear model predictive control, {\tt acados}, Crazyflie2.1, AI-deck.}

\end{abstract}

%\tableofcontents

%===========================================================
\section{Introduction}\label{sec:intro}
%===========================================================
Quadrotors have become increasingly popular in recent years due to their versatility and ability to perform various tasks such as search and rescue, aerial photography, mapping, and package delivery. However, achieving precise and aggressive trajectory tracking remains a major challenge in quadrotor control. {Aggressive trajectory tracking refers to the ability of unmanned aerial vehicles (UAVs) to perform high-speed and high-acceleration maneuvers in challenging environments. To achieve this researchers have proposed various control laws and methods that exploit the dynamics of quadrotor systems. These methods aim to accurately track the position, velocity, acceleration, jerk, snap, yaw angle, yaw rate, and yaw acceleration of the quadrotor~\cite{aggressive1,aggressive2}.} One key method to achieve precise and aggressive trajectory tracking in quadrotors is Nonlinear Model Predictive Control (MPC). NMPC is a control strategy that uses a model of the system dynamics to predict its future behavior and then optimizes a control policy based on that prediction. This allows the system to adjust its behavior in real-time based on sensor measurements, making it well-suited for tasks such as aggressive maneuvers and navigation in complex environments. However, the computational demands of NMPC can be a challenge for embedded systems such as nano quadrotors, e.g., Crazyflie2.1~\cite{crazyflie2.1}, that {have} limited onboard computational resources.

Recently, an open-source software for embedded NMPC called {\tt acados} is introduced~\cite{acados_doc}. It is designed to be efficient and user-friendly and allows for implementing {NMPC algorithms} on embedded systems. It provides various features such as real-time iterative solvers, automatic code generation, and the ability to handle multiple objective functions and constraints. This makes it a suitable tool for implementing NMPC on embedded systems {such as nano-quadrotors.}

% The dynamics of Crazyflie2.1 are modeled using a nonlinear state-space representation. The states of Crazyflie2.1 include its position, velocity, and attitude. The control inputs to Crazyflie2.1 include the thrust and the moments of the four motors. The constraints of Crazyflie2.1 include the maximum thrust and moments that can be generated by the motors, as well as the maximum angular velocity.

This paper proposes an aggressive trajectory tracking method for {nano-quadrotor, Crazyflie2.1,} using NMPC based on {\tt acados}. The proposed method utilizes Crazyflie's dynamics and constraints to generate a prediction of its future behavior. It then uses that prediction to optimize a control policy that will drive Crazyflie2.1 to follow a desired trajectory. The software package {\tt acados} enables the implementation of the NMPC algorithm on embedded systems, which is particularly important for Crazyflie2.1 due to its limited computational resources. The NPMC uses an optimization problem to determine the control inputs that will drive Crazyflie2.1 to follow a desired trajectory while satisfying the system's constraints. The optimization problem is solved in real-time using the full\_condensing\_HPIPM and SQP\_RTI solver provided by {\tt acados}. {The simulation and physical experimental results show that the proposed method can accurately track aggressive trajectories in dynamic environments. }

The paper is organized as follows: Section 2 provides the problem statement, and Section 3 presents the objectives and motivation. Section 4 briefly overviews related work in quadrotor control using embedded NMPC. Section 4 presents the proposed aggressive trajectory tracking methodology. Section 5 discusses the results of the simulation and real-world experiments. Finally, Section 6 concludes the paper and discusses future work.

%----------------------------
\subsection*{Problem Statement}
%----------------------------
{This research aims to develop a highly precise and efficient control method for the embedded nano quadrotor, Crazyflie2.1 that can accurately track a desired trajectory while considering the dynamics and constraints of the system. The proposed method should be able to handle aggressive maneuvers and navigate complex environments with high accuracy and fast response times in dynamic environments. To get aggressive trajectory tracking in dynamic environments, there are also some other challenges such as accurately modeling the dynamics of nano-quadrotor, constraint handling, and real-time implementation on embedded systems. To address these challenges, this research proposes to use NMPC as the control strategy and {\tt acados} toolbox to implement the NMPC algorithm on the embedded system of nano-quadrotor, Crazyflie2.1. }

%----------------------------
\subsection*{Objective and Motivation}
%----------------------------
The objective and motivation of this research are to develop a highly precise and efficient control method for Crazyflie2.1 that can accurately track a desired trajectory while considering the dynamics and constraints of the system. The proposed research aims to use {\tt acados} based NMPC on the embedded system of Crazyflie2.1.

This paper will fill the research gap on the implementation of {\tt acados} based embedded NMPC on the {nano-quadrotor,} Crazyflie2.1, for aggressive trajectory tracking. Previous research on quadrotor control has primarily focused on using Proportional-Integral-Derivative (PID) controllers, which are simple and widely used but may not be as precise or efficient as NMPC for aggressive trajectory tracking. Some studies have also used advanced controllers such as LQR, LQG, and nonlinear control methods, but they did not focus on using NMPC using {\tt acados} on nano-embedded platforms.

In the field of embedded NMPC, some studies have proposed the use of embedded MPC for quadrotor control, but they did not focus on small and lightweight quadrotors such as Crazyflie2.1, additionally, they did not use {\tt acados} as the implementation tool. This research aims to fill this gap by developing an aggressive trajectory tracking method for {embedded nano-quadrotor,} Crazyflie2.1 using {\tt acados} based NMPC and evaluating its performance through simulations and experiments. This research will provide valuable insights into using {\tt acados} based embedded NMPC.

%----------------------------
\subsection*{Related Work}
%----------------------------
This section puts our proposed works into context, focusing on the most related work. Several research has been conducted on NMPC for quadrotor control. Recently, acados-based NMPC has drawn much attention for embedded systems like quadrotors, thanks to the advances in hardware and algorithmic efficiency~\cite{bicego2018,foehn2021,kazim2022,romero2022}. Barbara Carlos et al.~\cite{carlos2020} present the design and implementation of an efficient position controller for quadrotors based on real-time NMPC with time-delay compensation and bounds enforcement on the actuators. In~\cite{torrente2021}, the authors presented gray-box gaussian process MPC in which aerodynamic effects are trained and modeled as Gaussian processes and incorporated into an MPC to achieve efficient and precise real-time feedback control, leading to up to $70\%$ reduction in trajectory tracking error at high speeds. Huan Nguyen et al. in their paper~\cite{nguyen2021} present a review of the design and application of MPC and NMPC control strategies for quadrotors. Furthermore, they present an overview of recent research trends on the combined application of modern deep reinforcement learning techniques and MPC for multi-rotor vehicles. Martin Saska and Tiago Nascimento described the embedded fast NMPC in~\cite{nascimento2021} to ensure the implementation of the position controller safely and stably for micro aerial vehicles that use low-processing power boards. Robin et al.~\cite{verschueren2018} introduced a new software package for embedded optimization called {\tt acados}, a new software package for MPC; we used this embedded optimization method to control Crazyflie2.1 by adjusting different parameters.
 
However, these existing studies did not focus on low-cost low-end embedded platforms. The proposed research aims to fill this gap by developing an aggressive trajectory tracking method for Crazyflie2.1 using {\tt acados} based NMPC and evaluating its performance through both simulations and hardware experiments.

%===========================================================
\section{Methodology}\label{sec:method}
%===========================================================

%----------------------------
\subsection{System Overview and Dynamics}
\begin{figure}[t]
	\centering
	\includegraphics[width=\textwidth]{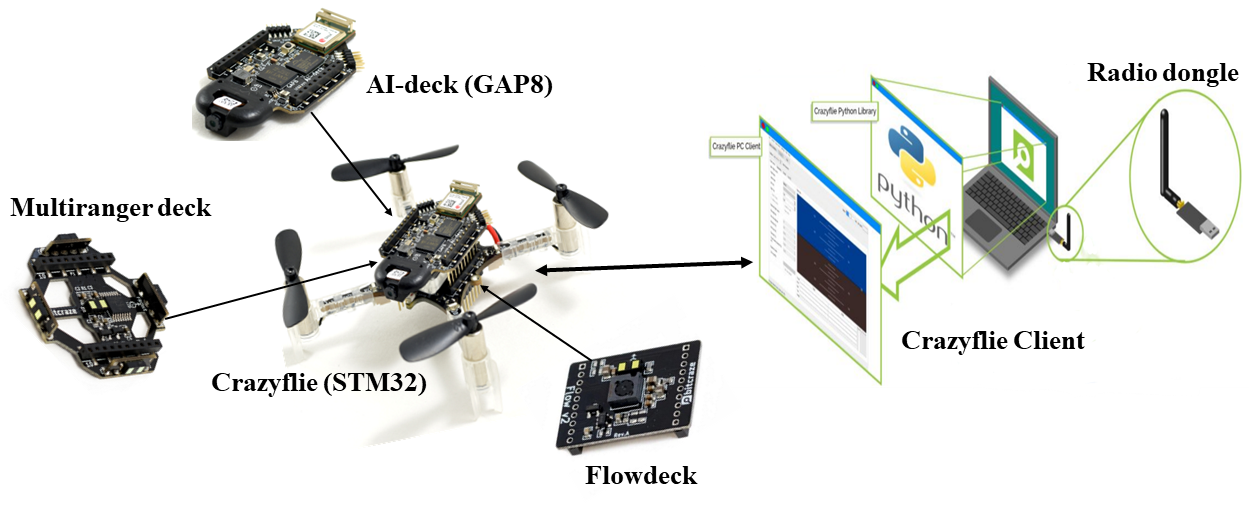}\vspace{-5mm}
	\caption{{Hardware and software configurations of the autonomous flying Crazyflie.}}
	\label{fig:system_overview}
\end{figure}

%----------------------------
{The nano-quadrotor has many components for autonomous flight in dynamic environments, Fig. \ref{fig:system_overview} shows the components and their connections to each other. The central part of the system is the Crazyflie with its MCU (STM32), where the autonomous flight is controlled, estimation of its position, collecting data from the sensor decks, and communication with other components. The Crazyflie is connected with two sensor expansion decks, the MultiRanger deck and the Flow deck. The MultiRanger deck detects any object around the Crazyflie while the Flow deck keeps track of the drone’s movements. The AI deck is also connected to the Crazyflie with its own MCU (GAP8), where the classification is run. It sends the classification result to Crazyflie, which relays the information along with the drone’s estimated position to the Crazyflie client’s console via radio.} 

{The proposed control architecture of nano quadrotor, Crazyflie2.1, is shown in Fig.~\ref{fig:crazyflie_control},} with a rigid body of mass $m$ and diagonal moment of inertia matrix $J=diag(J_x, J_y, J_z) {\in}  {\mathbb{R}^{3\times3}}$. The body-fixed frame \{B\} is located at the center-of-mass (COM) of a Crazyflie2.1 and aligned with a North-West-Up frame \{I\}. Then consider a nan0-quadrotor Crazyflie2.1 with position $p = (x, y, z)^{\top} \in {\mathbb{R}^3}$, expressed in \{I\}. attitude $q=(q_w, q_x, q_y, q_z) \in \mathbb{H}$, linear velocity $v_b = {(v_x, v_y, v_z)}^{\top} \in \mathbb{R}^3$ expressed in \{B\} and angular rate $w = {(w_x, w_y, w_z)}^{\top} \in \mathbb{R}^3$. The equations of motion of a Crazyflie2.1 in quaternion rotation matrix form are given by~\cite{erskine2021, sanwale2020}:

\begin{figure}[t]
	\centering
	\includegraphics[width=\textwidth]{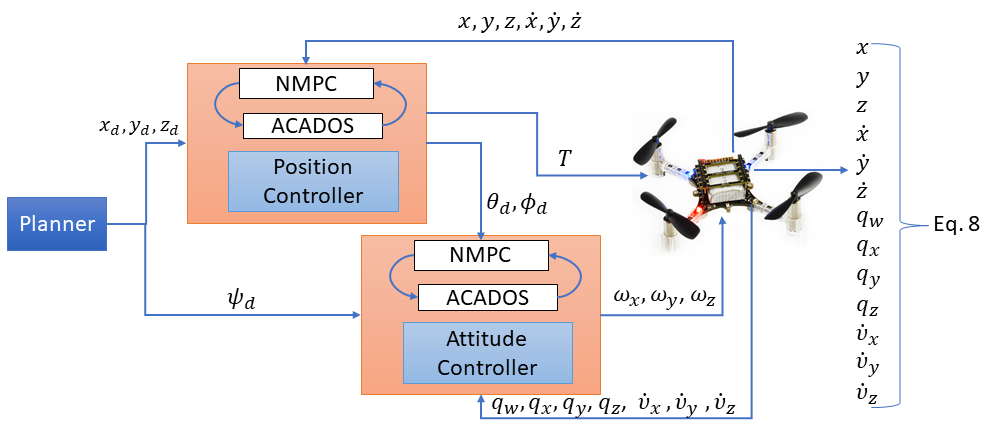}\vspace{-4mm}
	\caption{Proposed control architecture for an autonomous nano-quadrotor.}
	\label{fig:crazyflie_control}
        \vspace{-0.35cm}
\end{figure} 

% \vspace{-1.8cm}
\begin{equation}\label{quad_quaternion_model}
    \dot{\xi} = f(\xi, u) = \left\{
    \begin{array}{llll}				
    \dot{p} &= {v_b} \\
    \dot{v} &= 
    \frac{T_b}{m} 
    \begin{bmatrix}
        2 (q_w q_y + q_x q_z) \\
        2 (q_y q_z - q_w q_x) \\
        1 - 2(q_x^2 + q_y^2 ) \\
    \end{bmatrix} + g \\			
    \dot{q} & = \frac{1}{2}
    \begin{bmatrix}
        0 \\
        \omega \\
    \end{bmatrix} \otimes {q} \\
        \dot{\omega} &= {J}^{-1} (\tau_b - ({\omega} \times {J}{\omega})
    \end{array}
    \right.
\end{equation}
where $T$ is the thrust, and $\tau_b$ is the steering moments applied to the COM of the quadrotor Crazyflie2.1, with the state: $\xi:= {(p, q, v_b, w)}^{\top} \in \mathbb{R}^{13}$. A quaternion is a hypercomplex number of rank 4, and the quaternion-vector product is performed by the Kronecker product, denoted by $\otimes$, representing a rotation of the vector as in $ q \otimes v = q. {[0, v^{\top}]}^{\top} . \Bar{q}$, where $\Bar{q}$ is the quaternion's conjugate.

The nano-quadrotor, Crazyflie2.1, is controlled by setting the angular velocities of four co-planar propellers. Each propeller generates a thrust force $T_b = \Sigma_{i=1}^{4} k_t \Omega_{i}^2$, where $k_t$ is the aerodynamic coefficients. The vector of squared propeller velocities $[\Omega_{1}^2, \cdots , \Omega_{4}^2]^{\top}$ can be directly related to the actuation wrench $W_i = [{T_b} \ \  {\tau_b}^{\top}]^{\top}$ by
\begin{equation}
    W_i = {\Gamma}{\omega_i}
\end{equation}
where $\Gamma$ depends on the quadrotor geometry and on the aerodynamic coefficients of the propellers~\cite{identification2020}.

%----------------------------
\subsection{Embedded Numerical Optimal Control Using {\tt acados}}
%----------------------------
The NMPC controller is designed using a new software package for embedded optimization, called {\tt acados}~\cite{acados2022, acados_doc}. It is open-source software that provides a flexible and efficient framework for solving NMPC problems using sequential quadratic programming (SQP) \cite{boggs1995} and real-time iterations (RTI). It is implemented using CasADi \cite{andersson2019}, which enables automatic differentiation of the problem, and interfaces with high-performance linear algebra libraries, such as High-Performance Interior Point Method (HPIPM)~\cite{hpipm2020} and Basic Linear Algebra for Embedded Optimization (BLASFEO)~\cite{blasfeo2018}. This new software package aims to combine the objectives of flexibility, reproducibility, modularity, and efficiency. 
%{Fig. \ref{fig:acados_compare} shows the trade-off between sub-optimality and computation time in different solvers. It shows that acados and GRAMPC are faster and optimal as compared to other optimization solvers.}

%\begin{figure}[t]
%	\centering
%	\includegraphics[width=.8\textwidth]{5-diagrams/acados_compare.png}\vspace{-3mm}
%	\caption{{Trade-off between sub-optimality and computation time \cite{acados_doc}}.}
%	\label{fig:acados_compare}
%\end{figure}

{The two key components of {\tt acados} packages are the {\tt ACADOSOcpSolver} and {\tt ACADOSSimSolver}, these classes provide a flexible framework for solving optimal control problems and simulating nonlinear dynamic systems in low-cost embedded systems like Crazyflie2.1, AI-deck (GAP8). The AI-deck is equipped with a GAP8 system-on-chip processor from GreenWaves technology. GAP8 is a processor for the Internet of Things that enables low-cost performance. It is optimized for using a large spectrum of algorithms for images and audio. It allows the integration of artificial intelligence into devices that use the GAP8 processor. The main advantage of using the Greenwave chip is that it reduces deployment and operating costs.}

%----------------------------
\subsection{Nonlinear Model Predictive Control}
%----------------------------
{NMPC} is a feedback control algorithm that uses a system model to predict the system's future outputs and solves an optimization problem online to select an optimal control. In this paper, we designed the non-centralized NMPC structure, as it will reduce the computation time, so we divided the equation of motion of the {nano-quadrotor,} Crazyflie2.1 from (\ref{quad_quaternion_model}) into the translational and rotational models as:
\vspace{-0.35cm}
\begin{align}
    \text{Translational motion: }
    & \left\{
    \begin{array}{ll}				
    \dot{p} &= v \\
    \dot{v} &= 
    \frac{T_b}{m} 
    \begin{bmatrix}
        2 (q_w q_y + q_x q_z) \\
        2 (q_y q_z - q_w q_x) \\
        1 - 2(q_x^2 + q_y^2 ) \\
    \end{bmatrix} + {g} \\	
    \end{array}
    \right.
%\end{align}
%\begin{align}\label{rotational_model_quaternion}
\label{translation_model_quaternion}
\\ \nonumber \\
\text{Rotational motion: }
&\left\{
\begin{array}{ll}		
    \dot{q} & = \frac{1}{2}
    \begin{bmatrix}
        0 \\
        \omega \\
    \end{bmatrix} \otimes q \\
    \dot{\omega} &= {J}^{-1}({\tau}_b - {\omega}_i \times J {\omega}_i)
\end{array}
\right.
\label{rotational_model_quaternion}
\end{align}
In its most general form, NMPC solves an optimal control problem (OCP) by finding an input command $ u $ which minimizes a cost function $ J $ subject to its system dynamics model $ \dot{x} = f(x,u)$ while accounting for constraints on input and state variables for current and future time steps. To solve the aforementioned OCPs, we approximate by discretizing the underlying continuous-time OCPs and assuming linear least squares objectives using direct multiple shooting method~\cite{multiple1984}, which leads to the following nonlinear programming problem (NLP)~\cite{carlos2020}: 
\begin{equation}\label{mpc_optimization}
\begin{aligned}
            \min_{\xi,u} \quad & \frac{1}{2} \Sigma_{i=0}^{N-1} \|\eta(\xi_i, u_i) - \eta_i\|^2_{{W}} + \frac{1}{2}\|\eta_N(\xi_N) - \eta_N \|^2_{{W}_N} \\
            \textrm{s.t.} \quad & x_{i+1} = f(x_{i},u_{i}), \ i=0,1, \cdots, N-1 \\
                          \quad & u_{min} \leq u_{i} \leq u_{max}, \ i=0,1, \cdots, N-1 \\
        \end{aligned}
\end{equation}
where $\xi$ denotes the state vector defined as
\begin{equation}
    \xi = [x, y, z, q_w, q_x, q_y, q_z, v_x, v_y, v_z]^{\top} \in {\mathbb R}^{11}
\end{equation}
that is derived from the equations of motion (\ref{translation_model_quaternion}) and (\ref{rotational_model_quaternion}).
The control input $u =  [T, w_x, w_y, w_z]^{\top}$ is a concatenation of the total thrust and the three-axis angular velocities that are constrained as
\begin{equation}\label{control_input}
        T_{min} \leq T \leq T_{max}, \quad 
        -4 \pi \leq \omega_x, \omega_y, \omega_z \leq 4 \pi \\
\end{equation}

Using the translational (\ref{translation_model_quaternion}) and rotational (\ref{rotational_model_quaternion}) dynamics of the quadrotor Crazyflie2.1 in the quaternion coordinates, the state space equations are given by
\begin{equation}\label{states}
\begin{aligned}
    \dot{x} &= v_x , \ 
    \dot{y} = v_y , \ 
    \dot{z} = v_z, \\
    \dot{q}_w &=\frac{1}{2}( - \omega_x q_x - \omega_y q_y - \omega_z q_z) , \
    \dot{q}_x =\frac{1}{2}( \omega_x q_w + \omega_z q_y - \omega_y q_z), \\
    \dot{q}_y &=\frac{1}{2}( \omega_y q_w - \omega_z q_x + \omega_x q_z) , \
    \dot{q}_z =\frac{1}{2}( \omega_z q_w + \omega_y q_x - \omega_x q_y) , \\
    \dot{v}_x &\!=\!  2( q_w q_y + q_x q_z )\frac{T}{m},  \,
    \dot{v}_y \!=\!  2(q_y q_z - q_w q_x )\frac{T}{m},  \,
    \dot{v}_z \!=\!  ( 1 - 2 q_x^2 - 2 q_y^2 )\frac{T}{m} - g .
\end{aligned}
\end{equation}

The initial condition of the states of the quadrotor is set as:
\begin{equation}\label{initial_staes}
	{\xi}_0 = \begin{bmatrix}
		x_0 & y_0 & z_0 & q_{w_0} & q_{x_0} & q_{y_0} & q_{z_0} & v_{x_0} & v_{y_0} & v_{z_0}\\
	\end{bmatrix}^{\top}
\end{equation}

While the desired trajectory will be fed to the NMPC as a reference by using the future $N$ waypoints of the trajectory, which contain the desired states at each time instant:
\begin{equation}\label{desired_states}
	{\xi}_d = \begin{bmatrix}
		x_d & y_d & z_d & q_{w_d} & q_{x_d} & q_{y_d} & q_{z_d} & v_{x_d} & v_{y_d} & v_{z_d}\\
	\end{bmatrix}^{\top}
\end{equation}

Similarly, the desired control inputs ${u}_d$ can also be fed to the NMPC controller:
\begin{equation}\label{desired_inputs}
	{u}_d = \begin{bmatrix}
		T_d & \omega_{x_d} & \omega_{y_d} & \omega_{z_d} \\
	\end{bmatrix}^{\top}
\end{equation}

To implement the NMPC problem on Crazyflie2.1 platform using {\tt acados}, we need to define the system dynamics eq. (\ref{states}), the cost function eq. (\ref{mpc_optimization}), and constraints eq. (\ref{control_input}) in Python. We can then use {\tt acados} Python interface to solve the NMPC problem and generate the optimal control inputs. The NMPC is implemented using the open-source {\tt acados}, and the pseudo-code is given in Algorithm~\ref{alg:1}.

% \begin{algorithm}
% \caption{Pseudo-code for NMPC using Acados} 
%  \begin{itemize}
%      \item Define the dimensions and dynamics of the system eq.(\ref{states}). 
%      \item Define the acados model or problem \\
%            \ e.g. model = acados.Model()
%     \item Define the acados OCP: \\
%             OCP = acados.OCP() \\
%             OCP.model = model
%     \item Set Objective: eq. (\ref{mpc_optimization}) \\
%         Q = np.diag([]) \\
%         P = np.diag([]) \\
%         OCP.minimize(Q, R)
%     \item Set Constraints: eq. (\ref{control_input}) \\
%         $x_{min}$, $x_{max}$, $u_{min}$, $u_{max}$ \\
%         OCP.subject\_to($x_{min}$, $x_{max}$, $u_{min}$, $u_{max}$ )
%     \item Define Solver \\
%         Solver = acados.OCPSolver(OCP, Acados\_SIM)
%     \item Define the initial state and control: eq.(\ref{initial_staes}), eq. (\ref{desired_states}), eq. (\ref{desired_inputs})\\
%         $x_0$, $u_0$ = np.array([])
%     \item solves the problem \\
%         x, u, \_, \_ = solver.solve($x_0$, $u_0$)
%  \end{itemize}
% \end{algorithm}

\begin{algorithm}[t]
	\caption{Pseudo-code for NMPC using {\tt acados}}\label{alg:1}
	\begin{algorithmic}[1]
            \State \textbf{Define States:} $\xi$: States eq.(\ref{states}), $\xi_0$: Initial state eq.(\ref{initial_staes}), $\xi_d$: Desired state eq.(\ref{desired_states}).
            \State \textbf{Define Inputs:} $u$: control inputs eq.(\ref{control_input}), $u_d$: desired inputs eq.(\ref{desired_inputs}).
            \State \textbf{Set the constraints:} $u_{min} \leq u_{i} \leq u_{max}$: bounds on the control input.
		\State $N$: Prediction horizon.
		\For{\texttt{i in range}}
		\State Get the current state $\xi$ from eq. (\ref{states}).
		\State Update the current state $\xi$, desired state $\xi_d$, desired inputs $u_d$.
		\State set initial guess for optimization  $\xi_0$ as defined in eq. (\ref{initial_staes}).
		\State solve the optimization using {\tt acados} solver, 
            \State \quad $\min_{\xi,u} \frac{1}{2} \Sigma_{i=0}^{N-1} \|\eta(\xi_i, u_i) - \eta_i\|^2_{{W}} + \frac{1}{2}\|\eta_N(\xi_N) - \eta_N \|^2_{{W}_N}$. 
		\State Get optimal control input $u_i  =  [T, w_x, w_y, w_z]$. 
		\State simulate the system with the optimal control input $x_{i+1} = f(x_{i},u_{i})$.
		\EndFor
	\end{algorithmic}
\end{algorithm}

%===========================================================
\section{Simulation Results}\label{sec:sim}
%===========================================================
The simulation results of {nano-quadrotor} Crazyflie 2.1 using {\tt acados} based NMPC demonstrate significant improvements in precision and control. The implementation of NMPC enables the {nano-quadrotor} Crazyflie2.1 to achieve highly accurate position and attitude tracking, {in dynamic environments and model uncertainties. The NMPC controller is capable of adjusting the desired trajectory on-the-fly to ensure the quadrotor can handle the dynamic disturbances.} Additionally, {\tt acados} tool provides a fast and robust solution, allowing for real-time optimization and control of the nano-quadrotor's movements. The hovering and trajectory tracking simulation results highlight the effectiveness and potential of using {\tt acados} based NMPC for advanced control of {nano-quadrotor} Crazyflie 2.1.

%----------------------------
\subsection{Hovering}
\begin{figure}[t]
	\centering
	\includegraphics[width=0.6\textwidth]{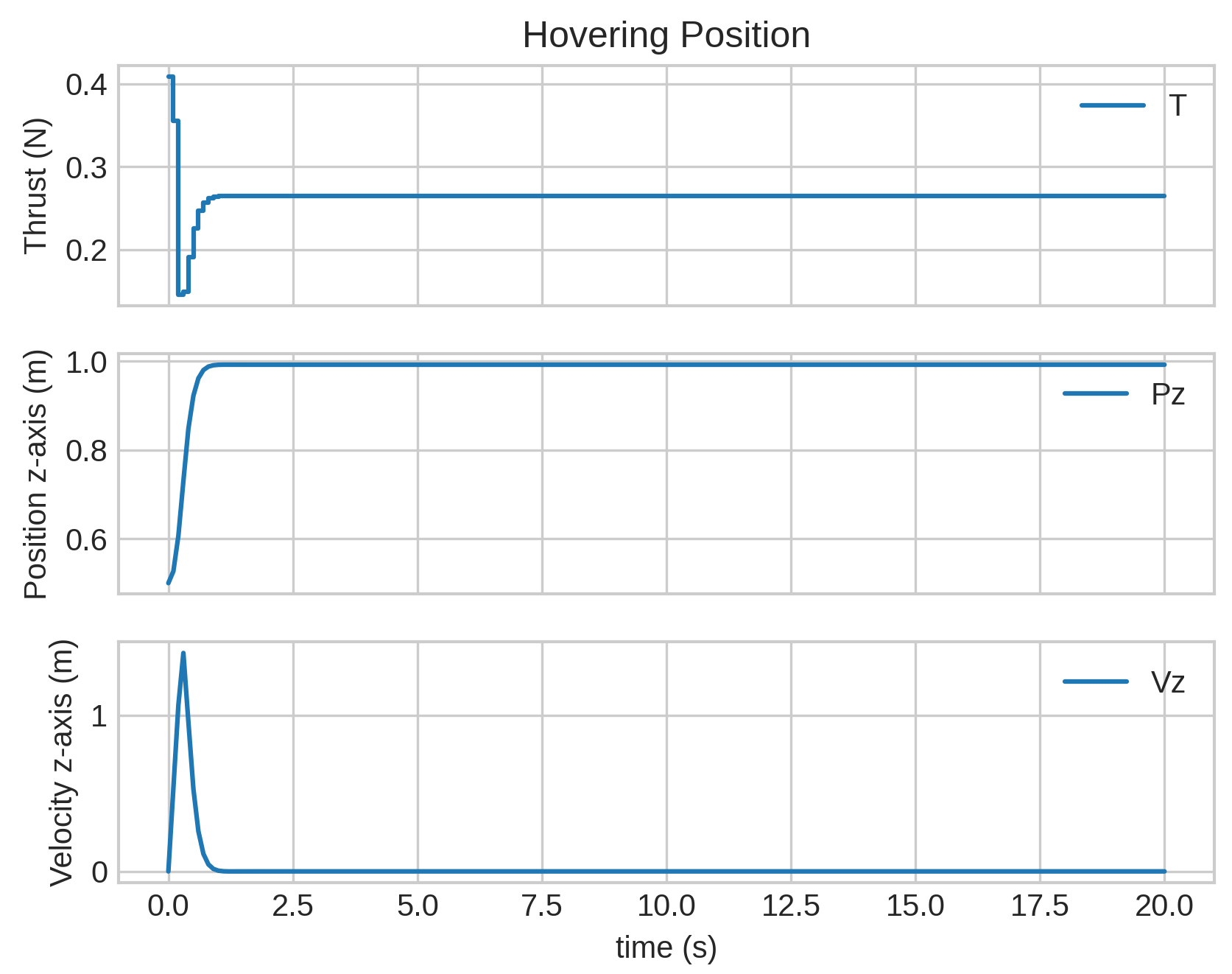}\vspace{-4mm}
	\caption{Hovering of Crazyflie2.1 at an altitude of 1.0 meters.}
	\label{fig:hovering_1}
\end{figure}

\begin{figure}[t]
	\centering
	\includegraphics[width=0.6\textwidth]{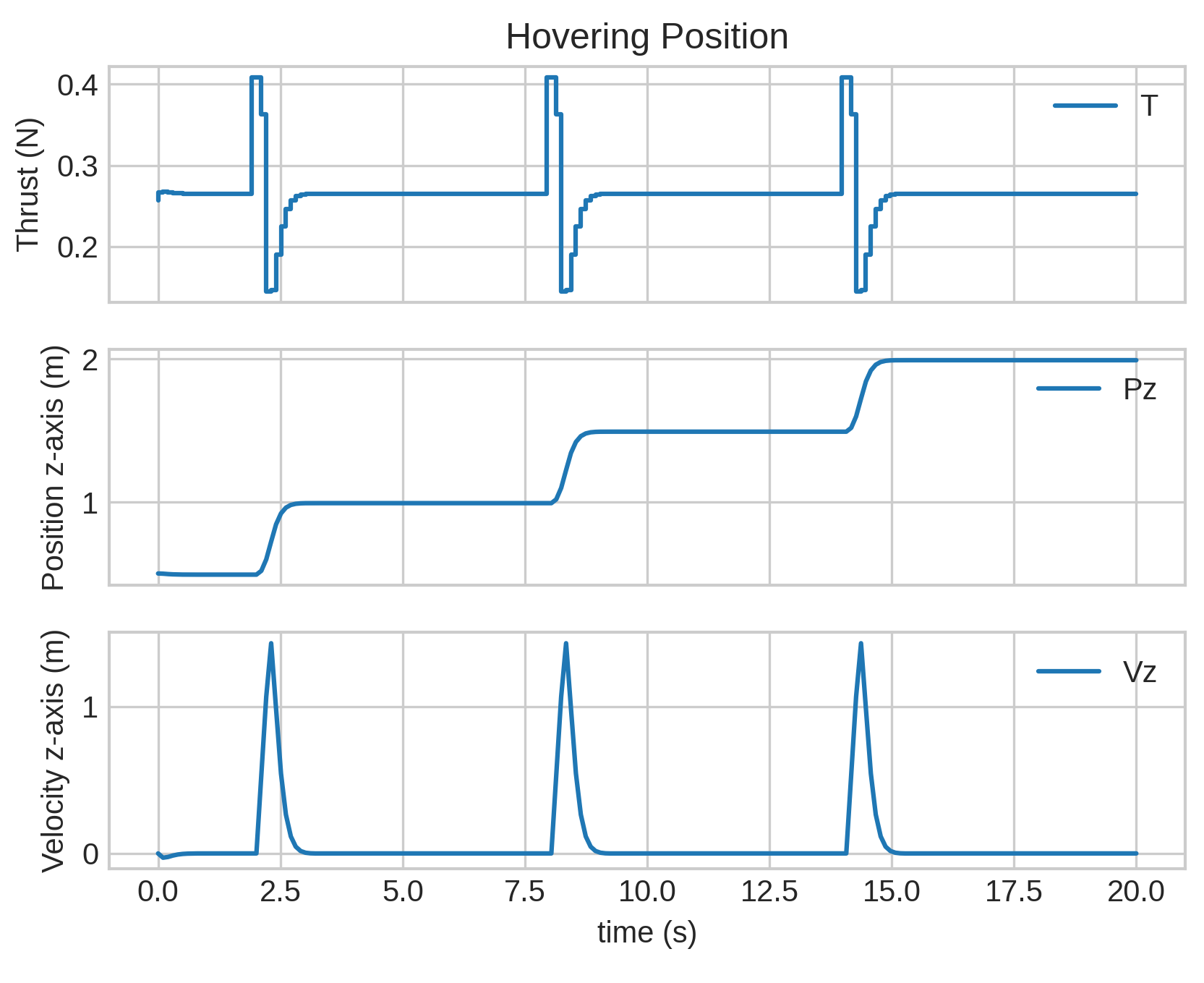}\vspace{-6mm}
	\caption{Hovering of Crazyflie2.1 at altitudes of 0.3, 1.0, 1.5, and 2.0 meters.}
	\label{fig:hovering_2}
\end{figure}

%----------------------------
The simulation results of {nano-quadrotor} Crazyflie 2.1 hovering using {\tt acados} based NMPC exhibit precise and stable hovering behavior. The simulation results in Fig. \ref{fig:hovering_1}, and Fig. \ref{fig:hovering_2} demonstrates that {\tt acados} based NMPC controller outperforms other existing methods and provides an effective solution for the hovering of quadrotor Crazyflie 2.1.  We select the prediction horizon $N = 10$ corresponding to the $t_f = 1$ CPU sec with the sampling-time $10$ Hz, and the total simulation time is $T = 20$ CPU sec. The precise and stable Hovering of Crazyflie2.1 at the position of $z = 1$ meter, the thrust input and velocity along the z-axis is shown in Fig. \ref{fig:hovering_1}. The average solver computation time is $0.00010268$ CPU sec while the solver maximum computation time is $0.000493765$ CPU sec. 

Fig. \ref{fig:hovering_2} shows the hovering of nano-quadrotor Crazyflie2.1 at different position steps; in the first step, Crazyflie2.1 stable at $z = 0.3m$ for 1 second and then flies towards $z = 1.0m$, stay there for 2 seconds, then hover at $z = 1.5m$ for 2 seconds and then finally hover at $z = 0.2m$ very precisely. To hover at different positions, the solver average computation time is $9.0456\times10^{-5}$ CPU sec, and the solver maximum computation time is $42.861\times10^{-5}$ CPU sec. 

%----------------------------
\subsection{Trajectory Tracking}
%----------------------------
The simulation results in Fig. \ref{fig: Trajectory_tracking_1} and Fig. \ref{fig: desired_trajectories} show promising performance of {nano-quadrotor} crazyflie2.1 trajectory tracking using {\tt acados} based NMPC. The quadrotor crazyflie2.1 successfully tracks a reference trajectory while accounting for model uncertainty. The NMPC controller effectively controls the position, velocity, and attitude of the quadrotor crazyflie2.1, maintaining its stability and trajectory tracking accuracy. The simulation results demonstrate the ability of {\tt acados} based NMPC to handle nonlinear and multivariable dynamics of crazyflie2.1. Here we demonstrate two case studies, takeoff cruise land, and helical trajectory tracking.

%~~~~~~~~~~~~~~
\subsubsection{Takeoff, Cruise, and Land}
\begin{figure}[t]
	\centering
	\includegraphics[width=0.6\textwidth]{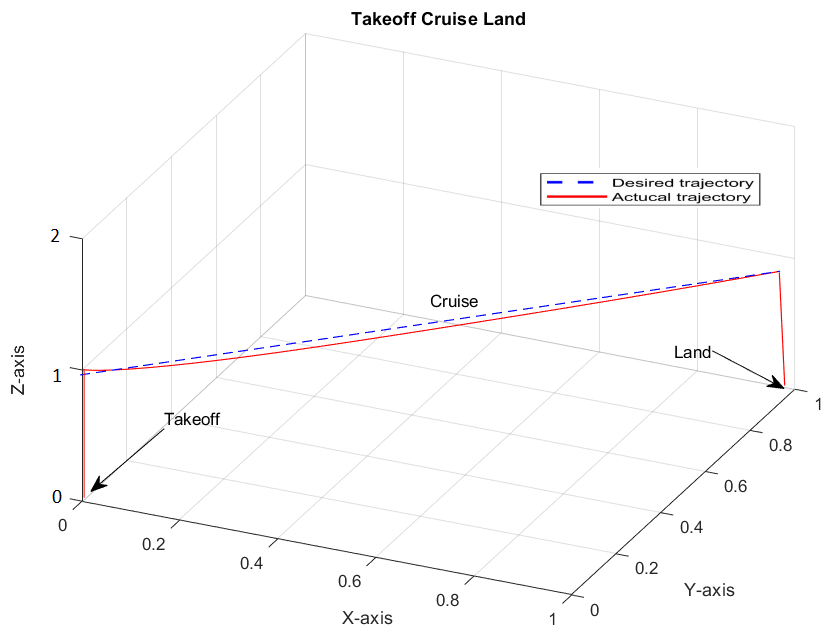}\vspace{-4mm}
	\caption{Trajectory tracking: Takeoff, Cruise and Land.}
	\label{fig: Trajectory_tracking_1}
\end{figure}

\begin{figure}[t]
	\centering
	\includegraphics[width=0.5\textwidth]{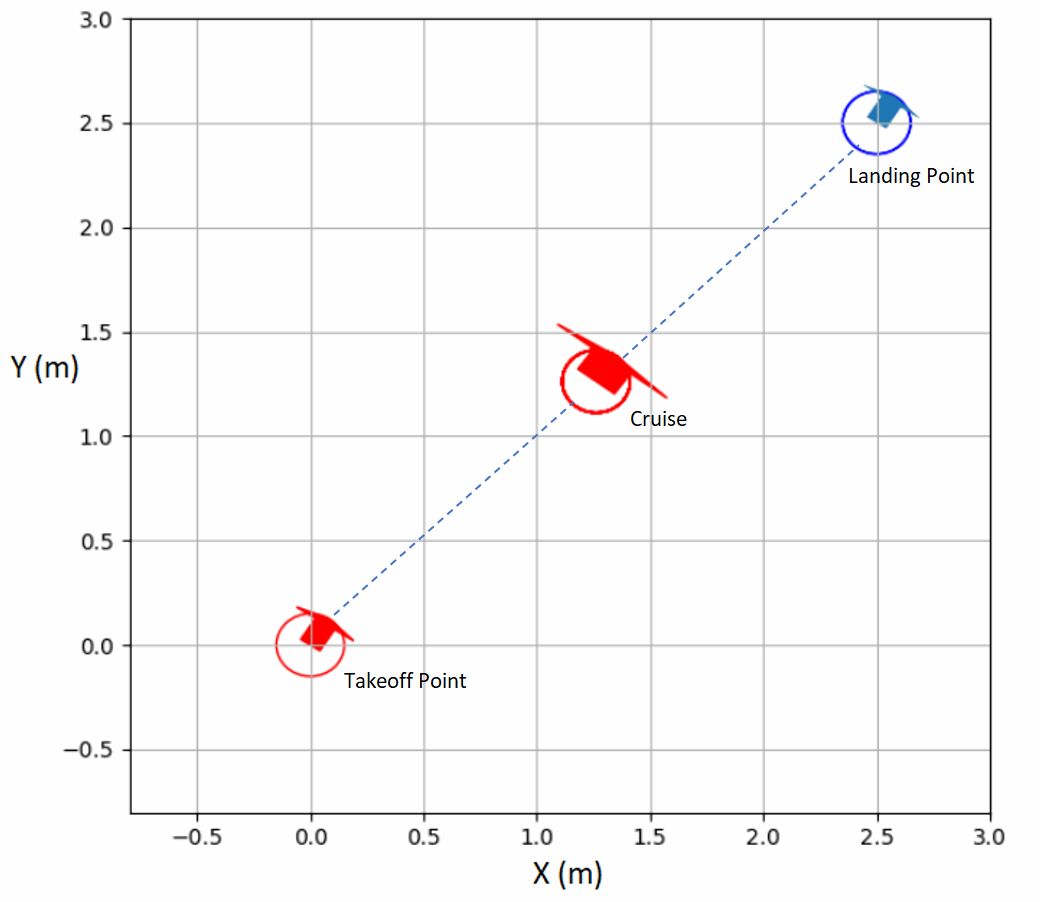}\vspace{-4mm}
	\caption{Trajectory tracking: Moving from a point (A) to a point (B). Video of experimental results available at \href{https://youtu.be/u2ILt5vZLK4}{https://youtu.be/u2ILt5vZLK4}.}
	\label{fig: Trajectory_tracking_2}
\end{figure}

%~~~~~~~~~~~~~~
The simulation results of crazyflie2.1 takeoff\_cruise\_land trajectory in Fig. \ref{fig: Trajectory_tracking_1} and Fig. \ref{fig: Trajectory_tracking_2} demonstrate the effectiveness of the control strategy in performing complex maneuvers using {\tt acados} based NMPC. During the takeoff phase, Crazyflie2.1 quickly reaches a stable hover state, while in the cruise phase, the controller maintains Crazyflie2.1 's altitude and heading while accounting for model uncertainties and external disturbances. In the landing phase, the controller accurately guides the quadrotor crazyflie2.1 toward the landing spot and ensures a smooth landing. The simulation results in Fig. \ref{fig: Trajectory_tracking_1} and Fig. \ref{fig: Trajectory_tracking_2} showcase the capabilities of {\tt acados} based NMPC in handling challenging maneuvers, including takeoff, cruising, and landing. In Fig. \ref{fig: Trajectory_tracking_2}, we also demonstrate the quadrotor Crazyflie2.1 cruise in animation form from point A to point B.  The average solver computation time for takeoff\_cruise\_land is $0.01587208s$ while the solver maximum computation time is $0.00968526s$.

%~~~~~~~~~~~~~~
\subsubsection{Helical Ascending Flight}
\begin{figure*}[t]
\centering
\subfloat[Position and velocity along $x$-axis.]{
    \includegraphics[width=0.49\textwidth]{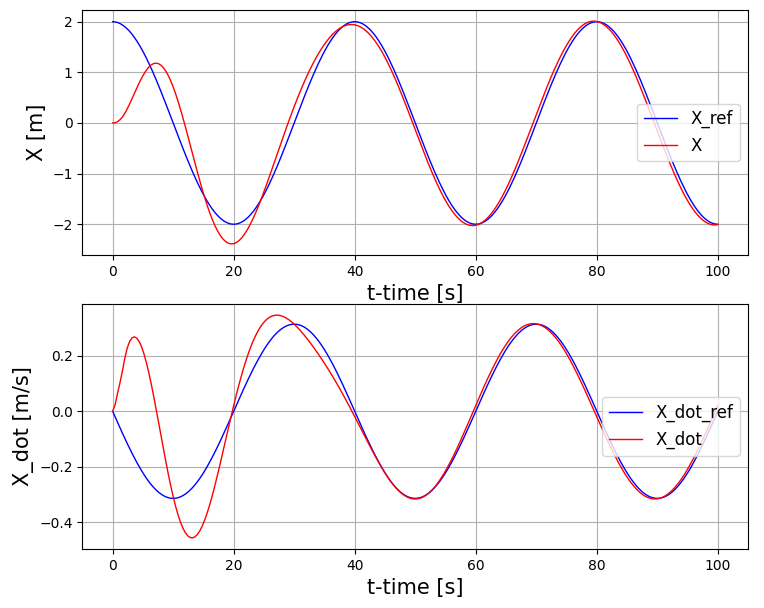}%
    \label{fig: Position_along_x}
}
\subfloat[Position and velocity along $y$-axis.]{
    \includegraphics[width=0.49\textwidth,height=.39\textwidth]{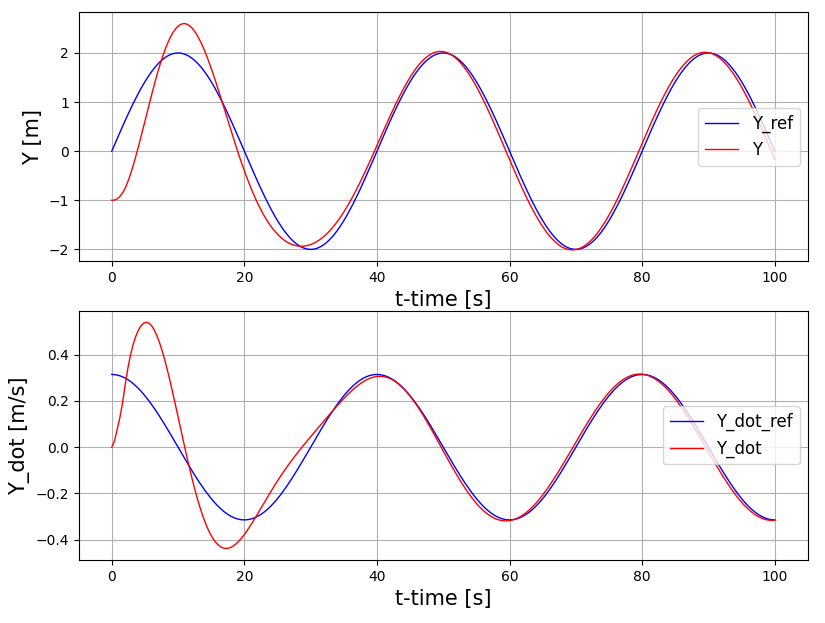}%
    \label{fig: Position_along_y}
}
\\
\subfloat[Position and velocity along $z$-axis.]{
    \includegraphics[width=0.49\textwidth]{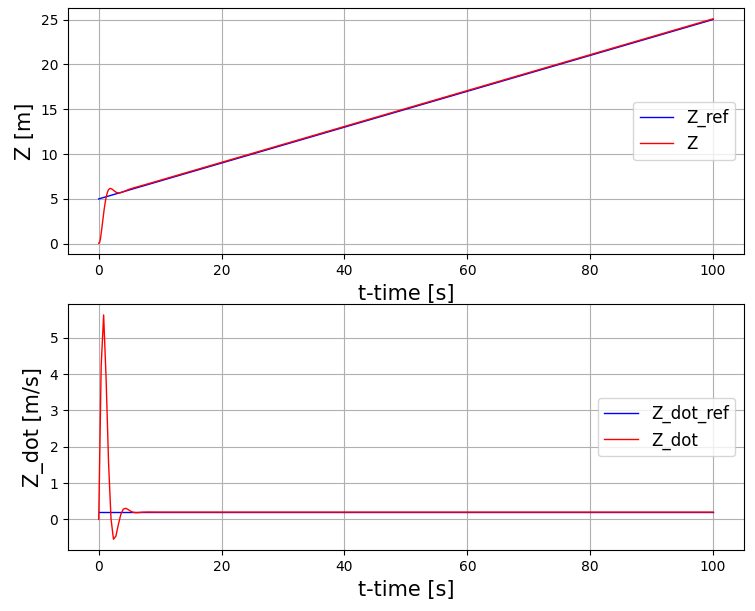}%
    \label{fig: Position_along_z}
}
\subfloat[Position and velocity along $y$-axis.]{
    \includegraphics[width=0.49\textwidth,height=.4\textwidth]{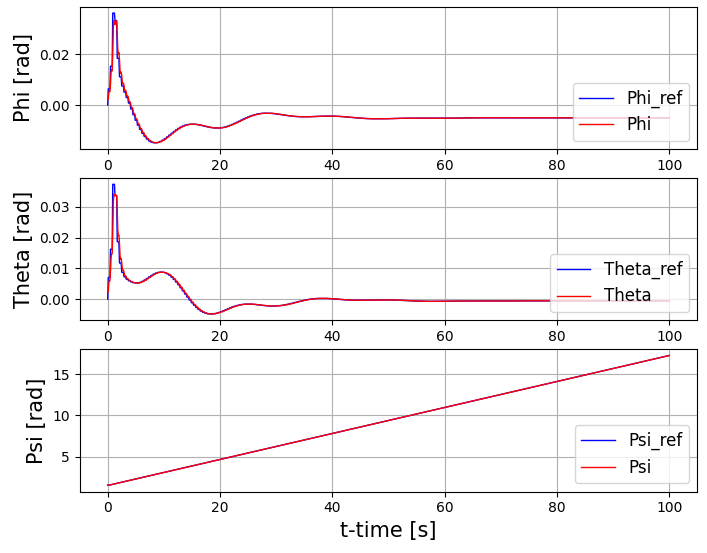}%
    \label{fig: angular positions}
}
  \caption{Position, velocity and orientation trajectories.}
\end{figure*}

\begin{figure}[t]
	\centering
	\includegraphics[width=.9\textwidth]{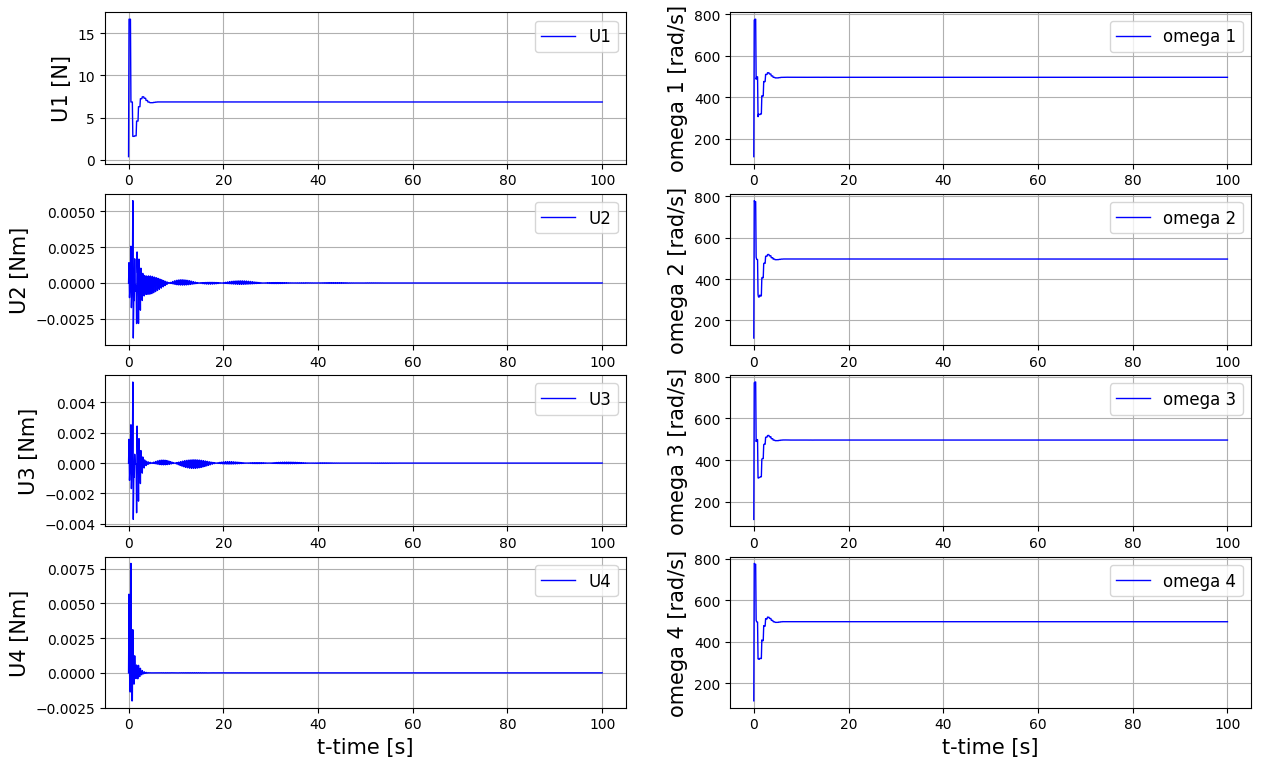}\vspace{-3mm}
	\caption{Input values and angular velocities as a function of time.}
	\label{fig: angular_velocities}
\end{figure}

\begin{figure}[t]
	\centering
	\includegraphics[width=0.45\textwidth]{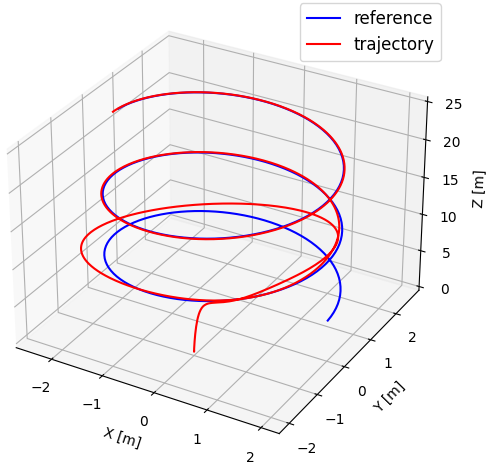}\vspace{-3mm}
	\caption{Reference (blue) and controlled (red) trajectory for the helical ascending flight simulation. A video of real hardware experimental results with Crazyflie2.1 is also available at~\url{https://youtu.be/j0Re-OuuWJc}.}
	\label{fig: desired_trajectories}
\end{figure}

%~~~~~~~~~~~~~~
The simulation results from Fig. \ref{fig: Position_along_x} to Fig. \ref{fig: desired_trajectories} demonstrate the ability of the control strategy to handle complex and dynamic trajectories using {\tt acados} based NMPC. The controller accurately tracks the helical trajectory, maintaining the quadrotor's orientation and position with high accuracy. Using {\tt acados} based NMPC allows for effective control of the quadrotor's attitude and altitude, accounting for model uncertainties and external disturbances.  
Fig. \ref{fig: Position_along_x} to Fig. \ref{fig: angular_velocities} shows time histories of flight trajectory, linear position, angular position, linear velocity, angular velocity, and input values to demonstrate the efficiency of proposed controllers for crazyflie2.1 quadrotor.

%\begin{figure}[t]
%	\centering
%	\includegraphics[width=0.475\textwidth]{4-Figures/Simulations/Trajectory_tracking/Halical/position_along_x_axis.png}\vspace{-4mm}
%	\caption{Position and velocity along $x$-axis}
%	\label{fig: Position_along_x}
%%\end{figure}
%%\begin{figure}[t]
%	\centering
%	\includegraphics[width=0.475\textwidth]{4-Figures/Simulations/Trajectory_tracking/Halical/Position_along_y_axis.png}\vspace{-4mm}
%	\caption{Position and velocity along $y$-axis}
%	\label{fig: Position_along_y}
%\end{figure}

%\begin{figure}[t]
%	\centering
%	\includegraphics[width=0.5\textwidth]{4-Figures/Simulations/Trajectory_tracking/Halical/position_along_z_axis.png}\vspace{-4mm}
%	\caption{Position and velocity along $z$-axis.}
%	\label{fig: Position_along_z}
%%\end{figure}
%%\begin{figure}[t]
%	\centering
%	\includegraphics[width=0.5\textwidth]{4-Figures/Simulations/Trajectory_tracking/Halical/angles_position.png}\vspace{-4mm}
%	\caption{Roll, Pitch, Yaw values as a function of time.}
%	\label{fig: angular positions}
%\end{figure}

The simulation result presented in Fig. \ref{fig: desired_trajectories} shows that the control strategy achieves a high success rate in tracking the reference trajectory with a small margin of error. However, tracking the x, y, and z reference velocity values exhibits significant overshoot at the start of the test period due to the quadrotor Crazyflie2.1 starting its journey far from the desired trajectory. Nonetheless, once the quadrotor Crazyflie2.1 arrive at the desired trajectory, their velocities stabilize and smoothly track the reference values. In Fig. \ref{fig: desired_trajectories}, we also demonstrate the quadrotor Crazyflie2.1 helical trajectory tracking in animation form.  The average solver computation time for helical trajectory tracking is $0.003206390s$ while the solver maximum computation time is $0.00597095$.

%===========================================================
\section{Experimental Results}\label{sec:exp}
%===========================================================

%----------------------------
\subsection{Setup}
%----------------------------
{To experimentally validate the proposed architecture, we utilized the Crazyflie 2.1 nano-quadrotor~\cite{crazyflie2.1} developed by Bitcraze, a Swedish company. The nano-quadrotor crazyflie is equipped with an AI-deck, a MultiRanger deck, and a Flow deck as shown in Fig. \ref{fig:system_overview}. The AI-deck has its own powerful microcontroller (GAP8), providing extra memory and computational resources. The MultiRanger deck detects obstacles, and the Flow deck tracks the drone's movement. To upload the code into AI-deck there are two methods one is through WiFi and the second one is through JTAB debugger. It starts with the python/C++ file (our code), then uses the GAP flow to use the model in the C file that controls the system. After that, it is flashed over to the AI-deck with the WiFi or JTAG debugger.}

%----------------------------
\subsection{Experiments Descriptions}
%----------------------------
To assess our proposed approach, nano-quadrotor, crazyflie2.1 executes three different trajectories (hover, takeoff\_cruise\_land, helical) as shown in the video1\footnote{\url{https://youtu.be/0_dN6XwYSYY}}. The parameters of crazyflie2.1 are given in table \ref{table_1}, while the setting {\tt acados} parameters, we choose full\_condensing\_HPIPM for qp\_solver, SQP\_RTI is the solver type, Gauss\_Newton is the Hessian approximation, in integrator\_type we used Explicit Runge\_kutta (ERK) method, while in simulation methods the number of stages is 4 and number of steps are 3, and maximum iteration for nonlinear programming iteration is selected as 200 and the tolerance is $1e^{-4}$. 

\begin{table}[t]
\centering
\caption{Parameters of Crazyflie2.1~\cite{crazyflie2.1}.}\label{table_1}
\begin{tabular}{l@{\,\,\,\,}c@{\,\,\,\,\,\,\,\,}l@{\,\,\,\,}l}
\hline
Parameters            & Symbol & Value                   & Units                  \\ \hline \hline
Mass                  & m      & 42                      & g                      \\
Prop-to-prop length   & l      & 92                      & mm                     \\
Inertial along x-axis & $I_x$  & $1.6571\times 10^{-5}$ & kg$\cdot$m$^2$         \\
Inertial along x-axis & $I_y$  & $1.6571\times 10^{-5}$ & kg$\cdot$m$^2$         \\
Inertial along x-axis & $I_z$  & $2.92\times 10^{-5}$   & kg$\cdot$m$^2$         \\
Thrust Coefficient    & c      & $2.88\times 10^{-8}$   & N$\cdot$m$^2$  \\
Drag Coefficient      & b      & $7.24\times 10^{-10}$   & Nm$\cdot$s$^2$   \\ \hline
\end{tabular}
\end{table}

The performance of the control strategy was evaluated in terms of its ability to track the desired trajectory while avoiding collisions with obstacles. The experimental results demonstrated the effectiveness of the NMPC control strategy in enabling precise trajectory tracking while navigating around {obstacles in dynamic environments.}

%===========================================================
\section{Conclusion and Future Work}
%===========================================================
This paper presents an aggressive trajectory tracking method for the {nano-quadrotor} Crazyflie2.1 using {\tt acados} based NMPC. {The proposed method can track the desired trajectories while considering the dynamics and constraints of Crazyflie2.1 in different dynamic environments.} Using {\tt acados} allows for efficient implementation of the NMPC algorithm on embedded systems {i.e., AI-deck,} making it well-suited for use on the limited computational resources of {nano-quadrotor} Crazyflie2.1. The simulation and the real-world test results show the proposed method's effectiveness in {trajectory tracking in different dynamic environments.} The proposed research work significantly contributes to the implementation of {\tt acados} based NMPC for nano-quadrotor Crazyflie2.1 by providing a new and efficient method for aggressive trajectory tracking. Some of the future work directions are the extension to multiple nano-quadrotor crazyflies2.1; the proposed method will be applied to other Unmanned Aerial Vehicles (UAVs) and Unmanned Ground Vehicles (UGVs), integration with other control techniques such as reinforcement learning and human-in-the-loop can be added to the control loop to increase the safety and reliability of the UAVs and UGVs. 

%===========================================================
\section*{Acknowledgments}
%===========================================================
This research was supported by the Basic Science Research Program through the National Research Foundation of Korea (NRF) funded by the Ministry of Education (NRF-2020R1F1A1076404, NRF-2022R1F1A1076260).

%----------------------------
\bibliographystyle{splncs03}
\bibliography{AggressiveTracking_EmbNMPC}
%----------------------------

\end{document}